\documentclass[runningheads]{llncs}
\usepackage{graphicx}
\usepackage{times}
\usepackage{marvosym}
\usepackage{amssymb}
\usepackage{latexsym}
\usepackage{graphicx}
\usepackage{amsmath}
\usepackage{booktabs}
\usepackage{multirow}
\newcommand{\citet}[1] {\citeauthor{#1}~\shortcite{#1}}
\usepackage[most]{tcolorbox}
\newcommand{\intuition}[1]{
\begin{tcolorbox}[colback=white,boxrule=1pt,top=0pt,bottom=0pt,left=1pt,right=2pt,top=2pt,bottom=2pt]
\em #1
\end{tcolorbox}
}

\begin{document}
\title{Kformer: Knowledge Injection in Transformer Feed-Forward Layers}
%
%
 \author{Yunzhi Yao\inst{1}\thanks{\; Contribution during internship at Microsoft Research.} \and
 Shaohan Huang\inst{2} \and
 Li Dong\inst{2} \and 
 Furu Wei\inst{2} \and
 Huajun Chen\inst{1} \and
 Ningyu Zhang\inst{1}$^{(\textrm{\Letter})}$}
 \authorrunning{Y. Yao et al.}
 \institute{Zhejiang University, Hangzhou, China \\
 \email{\{yyztodd,huajunsir,zhangningyu\}@zju.edu.cn}\\
 \and
 Microsoft Research, Beijing, China\\
 \email{\{shaohanh,donglixp,fuwei\}@microsoft.com}}
\maketitle              
\begin{abstract}
Recent days have witnessed a diverse set of knowledge injection models for pre-trained language models (PTMs); however, most previous studies neglect the PTMs' own ability with quantities of implicit knowledge stored in parameters. A recent study \cite{dai2021knowledge} has observed knowledge neurons in the  Feed Forward Network (FFN), which are responsible for expressing factual knowledge. In this work, we propose a simple model, Kformer, which takes advantage of the knowledge stored in PTMs and external knowledge via knowledge injection in Transformer FFN layers. Empirically results on two knowledge-intensive tasks, commonsense reasoning (i.e., SocialIQA) and medical question answering (i.e., MedQA-USMLE), demonstrate that Kformer can yield better performance than other knowledge injection technologies such as concatenation or attention-based injection. We think the proposed simple model and empirical findings may be helpful for the community to develop more powerful knowledge injection methods\footnote {Code available in \url{https://github.com/zjunlp/Kformer}.}.
\end{abstract}

\section{Introduction}

Pre-trained language models based on Transformer~\cite{DBLP:conf/nips/VaswaniSPUJGKP17} such as BERT~\cite{devlin-etal-2019-bert}, show significant performance on many Natural Language Processing (NLP) tasks.
However, recent studies have revealed that the performance of the knowledge-driven downstream task (for example, commonsense reasoning) is dependent on external knowledge; thus, the direct finetuning of pre-trained LMs yields suboptimal results.
\begin{figure}[ht]
   \centering
   \includegraphics[scale=0.5]{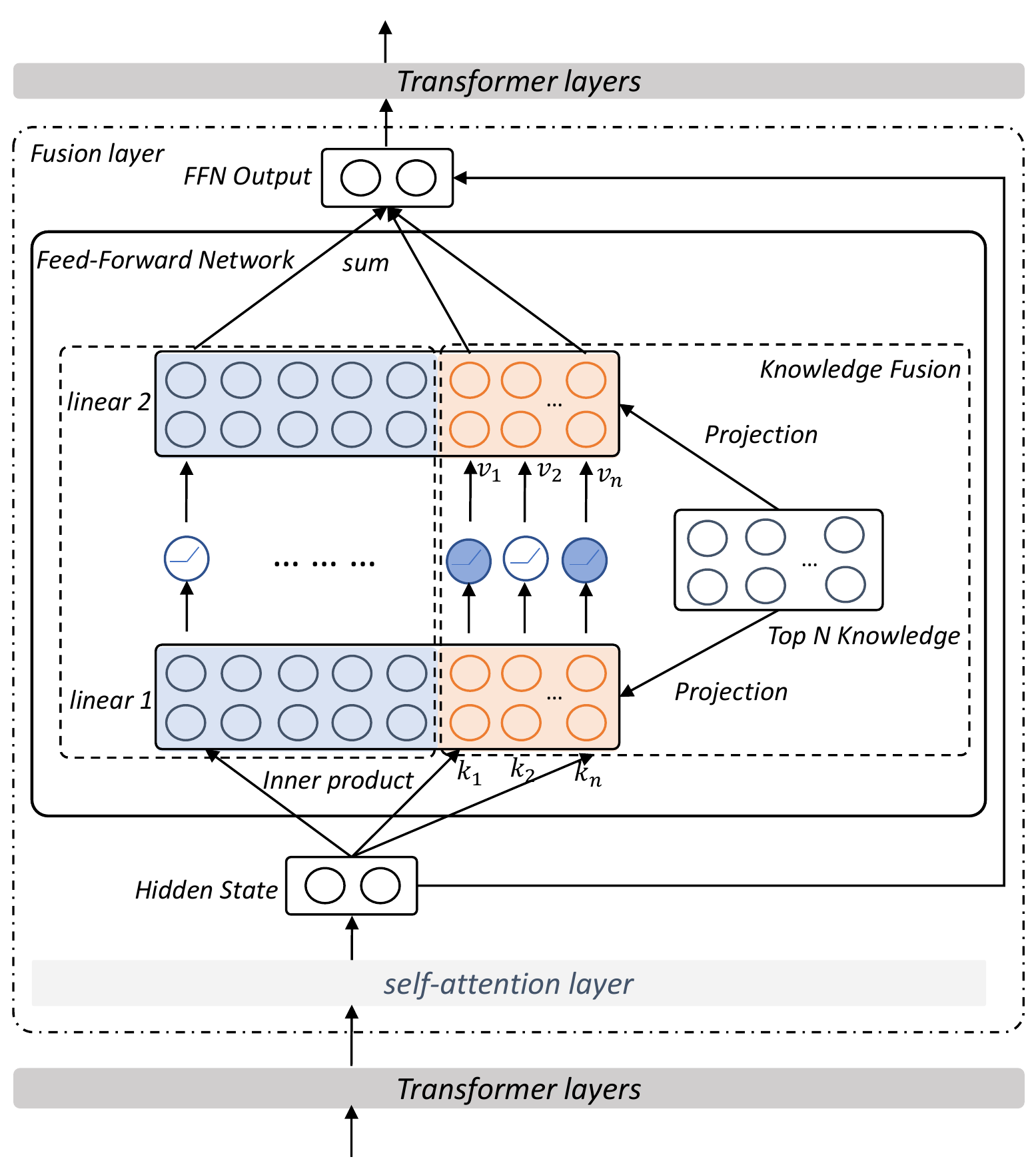}
   \caption{
   An illustration of KFormer for knowledge injection. 
   Input vectors are multiplied by each knowledge vector.
   The final output of the feed-forward network is the sum of the original linear layers' output and the knowledge fusion's output.}
   \label{fig:infuse}
\end{figure}

To address this issue, several works have attempted to integrate external knowledge into pre-trained LMs, which has shed light on promising directions for knowledge-driven tasks. 
On the one hand, several approaches ~\cite{xu2020fusing,mitra2020additional,lewis2020retrieval} try to fuse knowledge via concatenating the input sequence with external retrieved information (e.g., Wikipedia, knowledge fact).
On the other hand, other approaches ~\cite{DBLP:conf/emnlp/LinCCR19,chang-etal-2020-incorporating,wang2020connecting,Wang2019ImprovingNL,DBLP:conf/ijcai/ZhangDCCZZC21} obtain the representation of input and external knowledge separately and leverage interaction operations such as attention mechanism for knowledge fusion. 
However, most previous studies neglect the PTMs' own ability with quantities of implicit knowledge in parameters since a wealth of human knowledge is already captured by PTMs and stored in the form of ``modeledge'' \cite{han2021pre}. 

Recently, Dai~\cite{dai2021knowledge} has observed knowledge neurons in the  Feed Forward Network (FFN), which are responsible for expressing factual knowledge, which motivates us to investigate whether it is beneficial to inject external knowledge into the language model through FFN.
 
In this work, we try to take advantage of the \textbf{knowledge stored in PTMs and external knowledge} and propose a general knowledge fusion model named Kformer that injects knowledge in Transformer FFN layers.

As shown in Figure \ref{fig:infuse}, Kformer is a simple and pluggable model which can adapt to different knowledge types.
Specifically, we convert the knowledge into dense embedding vectors through a knowledge encoder and inject the knowledge into the Feed Forward Network of Transformer. 
We evaluate Kformer over several public benchmark datasets, including SocialIQA and MedQA. 
Experimental results show that Kformer can achieve absolute improvements in these tasks.
We further discover that it is complimentary for the PLMs to have more useful knowledge by knowledge injection through FFN.

In summary, we make the following contributions to this paper:
\begin{itemize}
\item We present a simple-yet-effective knowledge injection model Kformer to infuse the external knowledge into PLMs via Feed Forward Network.

\item Experimental results show the proposed appproach can obtain better performance and yield better interpretability.
\end{itemize}

\section{Knowledge Neurons in the FFN}

We first introduce some background of FFN and knowledge neurons.
Each layer in the Transformer contains a multi-head self-attention and a feed-forward network (FFN).
Primarily, the FFN consists of two Linear Networks. 
Suppose the final attention output of a Transformer layer is $\mathbf{x} \in \mathbb{R}^d$, the compute of the Feed Forward Network can be formulated as (bias terms are omitted):
\begin{equation}
\begin{array}{r}
 \mathrm{FFN}(\mathbf{x})=f\left(\mathbf{x} \cdot K^{\top}\right) \cdot V \\
    K, V \in \mathbb{R}^{d_{m} \times d}
\end{array}
\end{equation}
Here $K, V$ is the parameter matrices of the two Linear Networks. 
The input $\mathbf{x}$ are first multiplied by $K$ to produce the coefficients, which are then activated by $f$ and used to compute the weighted sum of $V$ as the feed-forward layer's output. 
Previous work~\cite{geva2020transformer} observes that FFN emulates neural memory and can be viewed as unnormalized Key-Value Memories.
More recently, Dai~\cite{dai2021knowledge} present preliminary studies on how factual knowledge is stored in pre-trained Transformers by introducing the concept of \textbf{knowledge neurons}. 
Note that the hidden state is fed into the first linear layer and activates knowledge neurons; then, the second linear layer integrates the corresponding memory
vectors. 

\intuition{\textbf{Motivating.}
Knowledge neurons in the FFN module are responsible for expressing factual knowledge; thus, it is intuitive to infuse knowledge by directly modifying the corresponding parameters in FFN, which can take advantage of the knowledge stored in PTMs and external knowledge.}

\section{Kformer: Knowledge Injection in FFN}
Kformer contains three main components: firstly, for each question, we retrieve the top $N$ potential knowledge from knowledge bases~(\S{\ref{sec:retrieval}}). 
Then, we obtain the knowledge representation via Knowledge Embedding~(\S{\ref{sec:embedding}}).
In the end, we fuse the retrieved $N$ knowledge into the pre-trained model via the feed-forward layer in Transformer (\S{\ref{sec:infuse}}).

\subsection{Knowledge Retrieval}
\label{sec:retrieval}
We first extract external knowledge for injection.
Here, we build a hybrid retrieval system to retrieve external knowledge.
Note that the retrieval process is not the main focus of this paper, and Kformer can be applied to different types of retrieval scenarios.
Specifically, we utilize an off-the-shelf sparse searcher based on Apache Lucene, Elasticsearch, using an inverted index lookup followed by BM25~\cite{robertson2009probabilistic} ranking. 
We select sentences with top $M$ scores from the results returned by the search engine as knowledge candidates. 
Since the IR system focuses primarily on lexical and semantic matching, we then conduct a simple dense retrieval on the $M$ candidates.
We obtain the dense representation of each knowledge via knowledge embedding (\S{\ref{sec:embedding}}) and compute the input sentence representation by the average token embeddings.
Then we calculate the score of each knowledge candidate by conducting an inner product on the input embedding and the knowledge embedding.
We choose knowledge candidates with the top $N$ scores as the knowledge evidence for knowledge injection (\S{\ref{sec:infuse}}).

\subsection{Knowledge Embedding}
\label{sec:embedding}
We view each knowledge candidate as a text sequence and leverage an Embedding layer to represent the knowledge. 
Each knowledge $k$ is firstly tokenized as $l$ tokens. 
Those tokens then would be embedded as $k_1, k_2, ..., k_l$ by the Knowledge Embedding. 
We initialize the knowledge embedding matrix as the same embedding matrix (first layer of the PTM) in the input sequence and update the Knowledge Embedding Layer simultaneously during training. 
The external knowledge is then represented as the average of these tokens' embedding: 

\begin{equation}
\mathrm{Embed}(k) = Avg (k_1,k_2,..,k_l)
\end{equation}

\subsection{Knowledge Injection}
\label{sec:infuse}
As shown in Figure \ref{fig:infuse}, Kformer injects knowledge in the Transformer FFN layer with the knowledge embedding. 
The feed-forward network in each Transformer layer consists of two linear transformations with a GeLU activation function. 
Suppose the final attention output of the layer $l$ is $H^l$, formally we have the output of the two linear layers as:
\begin{eqnarray}
    FFN(H^l) = f(H^l \cdot K^l) \cdot V^l
\end{eqnarray}
$K, V \in \mathbb{R}^{d_{m} \times d} $ are parameter matrices of the first and second linear layers and $f$ represents the non-linearity function.
$d_{m}$ refers to the intermediate size of Transformer and $d$ is the hidden size.
Suppose we acquire the top $N$ knowledge documents $\mathbf{k} \in \mathbb{R}^{d_{n} \times d}$ after retrieval~(\S{\ref{sec:retrieval}}). 
Through Knowledge Embedding, we can obtain each knowledge as $\mathbf{k_1},\mathbf{k_2},...,\mathbf{k_N} \in \mathbb{R}^{d}$. 
To inject the knowledge into the specific layer $l$, we need to map the knowledge to the corresponding vector space.
Here, for each layer $l$, we use two different linear layers to project the knowledge ($\Pr$ in the equation). 
$W_k^l$ and $W_v^l$ represents the weights of the two linear layers($W_k^l, W_v^l \in \mathbb{R}^{d \times d}$).
The two matrices $W_k^l$ and $W_v^l$ are initialized randomly and will be updated during fine-tuning.
\begin{eqnarray}
    \boldsymbol{\phi}_k^l = {\Pr}_k \mathbf{k} = W_k^l \cdot \mathbf{k} \\
    \boldsymbol{\phi}_v^l = {\Pr}_v \mathbf{k} = W_v^l \cdot \mathbf{k} 
\end{eqnarray}
After projection, we inject $\boldsymbol{\phi}_k^l$ and $\boldsymbol{\phi}_v^l$ into the corresponding $\boldsymbol{K}^l$ and $\boldsymbol{V}^l$. 
We expand the FFN by concatenating the projected knowledge to the end of the Linear layer and obtain the expanded $\boldsymbol{K}_E^l, \boldsymbol{V}_E^l \in \mathbb{R}^{(d_{m}+d_{n}) \times d}$.
Hence, after injection, the computation of FFN can be described as:  
\begin{equation}
\begin{array}{r}
FFN(H^l) = f(H^l \cdot \boldsymbol{K}_E^l) \cdot \boldsymbol{V}_E^l\\
= f(H^l \cdot [\boldsymbol{\phi}_k^l:\boldsymbol{K}^l]) \cdot [\boldsymbol{\phi}_v^l:\boldsymbol{V}^l]
\end{array}
\end{equation}

The model activates knowledge related to the input sequence and infuses knowledge with the query through the knowledge fusion component.
Next, the collected information will be processed and aggregated by the following transformer layer.
We simply inject the knowledge into the top 3 layers of the model and we analyze
the impact of injection layers in Section~\ref{sec:layer}.

\begin{table}[ht]
\centering
\caption{Overall statistics of the two datasets. 
`Know len.' means the length of knowledge text we extract from the knowledge Graph and  TextBook.
The question len. of the Social IQA dataset contains the length of the context.}
\scalebox{0.9}{
\begin{tabular}{lcc}
\bottomrule
\textbf{Metric}         & \textbf{Social IQA}   & \textbf{MedQA-USMLE}   \\ \hline
\# of options per question & 3                   & 4       \\
Avg. option len.           & 6.12                & 3.5     \\
Avg. question len.         & 20.16               & 116.6    \\
Avg./Max. know len.   & 8.63/37             & 55/1234 \\ \hline
\# of questions            &                     &         \\
Train                      & 33,410               & 10,178   \\
Dev                        & 1,954                & 1,272    \\
Test                       & 2,224                & 12,723   \\ \bottomrule
\end{tabular}}
\label{table:data}
\end{table}

\section{Experiments}

We do experiments on two multiple-choice tasks (Social IQA and MedQA-USMLE) to explore Kformer's performance on downstream tasks. We choose triples in the knowledge graph and documents in textbooks to examine the influence of different knowledge types.

\subsection{Dataset}

Social IQA~\cite{sap-etal-2019-social} is a commonsense QA task that is partially derived from an external knowledge source ATOMIC~\cite{sap2019atomic}. We use triples in ATOMIC as our knowledge candidates. For each triple, we paraphrase it into a language sequence according to the relation template in ATOMIC. 
MedQA~\cite{jin2020disease} is an OpenQA task in the biomedical domain, where questions are from medical board exams in the US, Mainland China, and Taiwan. We use the English dataset USMLE in our experiment. We adopt the textbooks provided by \cite{jin2020disease} as the knowledge base.

\subsection{Experiment Setting}

We summarize the statistics of each dataset in Table~\ref{table:data}, and we use accuracy as the evaluation metric for the two datasets.
Compared with SocialIQA, the lengths of the knowledge texts we used in MedQA is longer.

We detail the training procedures and hyperparameters for each of the datasets. 
We use RoBERTa as the backbone of Kformer and implement our method using Facebook's tool fairseq~\cite{ott2019fairseq}. 
We utilize Pytorch to conduct experiments with one NVIDIA RTX 3090 GPU.
All optimizations are performed with the AdamW optimizer with a warmup of learning rate over the first 150 steps, then polynomial decay over the remainder of the training.
Gradients were clipped if their norm exceeded 1.0, and weight decay on all non-bias parameters was set to 1e-2.
We mainly compare Kformer with the following injecting methods:

\begin{table}[ht]
\centering
\caption{Results of the Social IQA task. We compare our model with other models using ATOMIC as the knowledge source. The results marked with $^\dagger$ are retrieved from the leaderboard\protect\footnotemark ~of the dataset. }
\begin{tabular}{llcc}
\bottomrule
 & \textbf{Methods}           & \textbf{Dev Acc}     & \textbf{Test Acc}    \\ \hline
\multirow{4}{*}{Base Model} & RoBERTa                    & 72.26                & 69.26                \\
    & RoBERTa+MCQueen $^\dagger$      & -            & 67.22                   \\
   & RoBERTa+ATT & 72.10 & - \\
   & Kformer                    & 72.82                & 71.10                \\ \hline
\multirow{4}{*}{Large Model}                  & RoBERTa                    & 78.86                & 77.12                \\
& RoBERTa+MCQueen $^\dagger$ & \textbf{79.53}       & 78.00                \\ 
 & RoBERTa+ATT & 78.54 & - \\
 & Kformer                    & 79.27                & \textbf{78.68}       \\ \hline
\end{tabular}
\label{table:social_result}
\end{table}
\footnotetext{https://leaderboard.allenai.org/socialiqa/submissions/public}

\begin{table}[ht]
\centering
\caption{Results of the MedQA-USMLE task.}
\begin{tabular}{lcc}
\bottomrule
\textbf{Method}             & \textbf{Dev Acc} & \textbf{Test Acc} \\ \hline
IR-ES                        & 28.70    &  26.00    \\ \hline
MAX-OUT                      & 28.90    &  28.60    \\ \hline
RoBERTa               & 31.28    &  27.41    \\
RoBERTa+Concat         & 31.76    &  29.77    \\
RoBERTa+ATT           & 31.70  &  27.37\\
Kformer   & \textbf{33.02}    &  \textbf{30.01}    \\
\bottomrule
\end{tabular}
\label{table:med_result}
\end{table}

\paragraph{Concat:} 
This way is to simply concatenate the retrieved knowledge with the question as to the input of the model.
All the retrieved knowledge is joined together to make a single knowledge text $K$. 
The sequence of tokens {[CLS] $K$ [SEP] $Q$ [SEP] $A$} ($Q$ is the question, $A$ is the answer choice) is then passed to the model to get the final prediction. 
We denote this method as RoBERTa+Concat in the following part.

\paragraph{Attention:}
Multi-head self-attention plays a role of message passing between tokens \cite{Hao2021SelfAttentionAI}.
Input tokens interact with each other and determine what they should pay more attention to. We conduct experiments on injecting the knowledge into the self-attention module, the fusion is calculated as follows: 
\begin{equation}
\text {Attention}^{l}=\operatorname{softmax}\left(\frac{Q^{l}\left[\phi_{k}^{l} ; \boldsymbol{K}^{l}\right]^{T}}{\sqrt{d}}\right)\left[\phi_{v}^{l} ; \boldsymbol{V}^{l}\right]
\end{equation}

The knowledge is concatenated at the $\boldsymbol{K}$ and $\boldsymbol{V}$ parts in self-attention. 
During inference, the query will communicate with the injected knowledge.
We denote this method as RoBERTa+ATT in the following part.

\subsection{Experiments Results}

\begin{figure}[ht]
   \centering
   \includegraphics[scale=0.5]{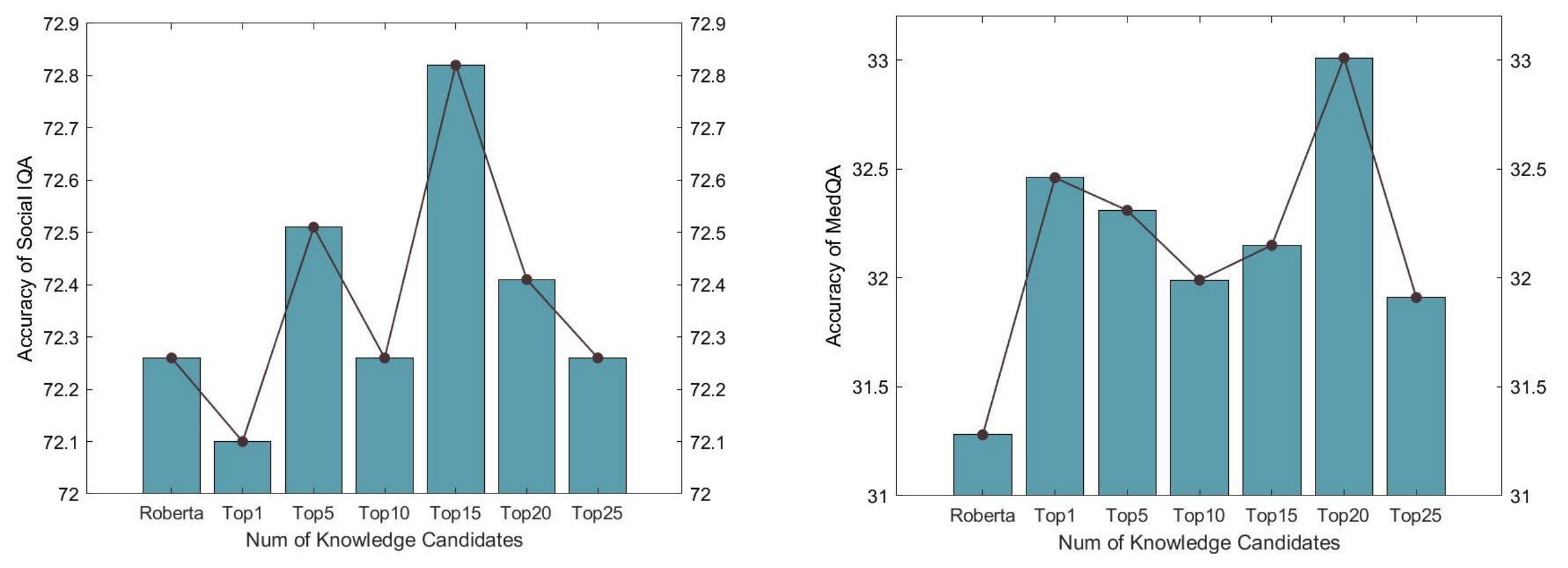}
   \caption{Results of the model with different top N (1, 5, 10, 15, 20, 25) retrieved knowledge candidates.}
   \label{analyze:quant}
\end{figure}

We list the result of the Social IQA in Table~\ref{table:social_result}. RoBERTa+MCQueen \cite{mitra2020additional} explored different ways to infuse knowledge into the language model, including concat, parallel-max, simple sum and weighted sum. 
All four methods put the knowledge into the model at the input part. 
We can observe from the table that adding the knowledge into the feed-forward layer can help the model learn the external knowledge, leading to 0.41 gain in the valid set and 1.56 gain in the test set for the large model. As compared to other methods of incorporating information into the model, our model has a distinct benefit. On the test set, our model outperforms the RoBERTa+MCQueen by 0.68.

\section{Analysis}

For the MedQA, we list the results in Table~\ref{table:med_result}. 
IR-ES adopts an off-the-shelf retrieval system based on ElasticSearch. 
For each question $q$ and each answer $a_i$, we get the score for the top N retrieved sentences and the option with the highest score is selected. 
We use the code provided by \cite{jin2020disease}\footnote{https://github.com/jind11/MedQA} to get the result. 
We compared our model with the concatenate method.

From the Table, we notice that Kformer is still competitive for knowledge with a long sequence length.
The model adding knowledge via the feed-forward layer outperforms the model concatenating the information into the input by 1.3 in the valid set and 0.3 in the test set. The average length of each knowledge text is 55 in the Textbook materials. When concatenating knowledge at the input, we may only focus on about ten potential pieces of evidence, while in our model, we can focus on much more evidence via injecting knowledge via FFN.

Meanwhile, we can find that on both two tasks, Kformer outperforms RoBERTa+ATT, indicating that injecting knowledge in FFN is better than injecting in attention.

\subsection{Impact of Top N Knowledge}
In this section, we examine the impact of the number of knowledge candidates retrieved in the knowledge retrieval part. We adjust the number of knowledge candidates and list the results in Figure~\ref{analyze:quant}.

\begin{table}[ht]
\centering
\caption{Results of the model with different knowledge infusion layer. 
We compute every three layers of Kformer and use the dev set of Social IQA task and dev and test set of the MedQA task.}
\begin{tabular}{lccl}
\bottomrule
\multirow{2}{*}{} & \textbf{Social IQA}  & \multicolumn{2}{c}{\textbf{MedQA}}       \\ \cline{2-4} 
                  & \multicolumn{1}{c}{\textbf{Dev}} & \multicolumn{1}{c}{\textbf{Dev}} & \textbf{Test}  \\ \hline
RoBERTa     & 72.26                   & 31.28                   & 27.41 \\ \hline
All layers       & 72.56                   & 32.31                   &  28.27  \\ 
Layer 10-12     & \textbf{72.82}                 & 33.02                   & 30.01 \\ 
Layer 7-9       & 72.10                   & \textbf{34.04}          & \textbf{30.32} \\ 
Layer 4-6       & 72.15                   & 31.21                   &  27.72  \\ 
Layer 1-3       & 72.51                   & 31.68                   &  29.30     \\ \bottomrule
\end{tabular}
\label{table:layers}
\end{table}

As shown in the Figure, the extracted information has both positive and negative effects on the model. If we retrieve less information, the model will be unable to solve the problem due to the lack of sufficient evidence. For the MedQA task, when retrieving the top 20 documents, Kformer performs best, better than retrieving the top 5, 10, and 15 documents. For the Social IQA dataset, retrieving the top 15 sentences achieves the best result. However, if we retrieve too much information for the question, the model will suffer from knowledge noise and make incorrect predictions. In the Figure, the performance of Social IQA decreases when retrieving information more than the top 15 sentences. The same pattern can be seen in the MedQA task; when retrieving more than the top 20 documents, the model's performance drops too.

\subsection{Impact of Layers}
\label{sec:layer}
In Kformer, the layer where we incorporate external knowledge is a critical hyper-parameter since different levels in Transformer may catch different information.
\cite{geva2020transformer} demonstrates that higher layer's feed-forward network learns more semantic knowledge. Meanwhile, \cite{wang-tu-2020-rethinking} shows that component closer to the model's input and output are more important than component in other layers. Here, we insert the knowledge at different levels in Transformer to see where it is the most beneficial. We conduct experiments on every three layers of the base model as the knowledge fusion layers. We also try to import knowledge to all of the Transformer layers. We list the results of
various knowledge infusion layers in Table~\ref{table:layers}.

On both tasks, the model that integrates evidence through the top three layers or the bottom three layers benefits from external knowledge, whereas applying knowledge into the model's 4-6 layers degrades its performance. It confirms \cite{wang-tu-2020-rethinking}'s finding that components closer to the model's input and output are more critical. Meanwhile, top layers typically outperform bottom ones, implying a higher layer's feed-forward network has gained more semantic information.
On the two tasks, layer 7-9 displays a different pattern. In the Social IQA task, the model that incorporates information via layers 7-9 performs the worst and degrades the model's performance, whereas in the MedQA task, adding knowledge via layers 7 to 9 performs the best. 
Finally, if we infuse knowledge into all layers, the model still has gained, but the result is lower than the top three layers' result, which could be owing to the detrimental effect of the intermediate layers.

\begin{figure*}[ht]
    \centering
   \includegraphics[scale=0.45]{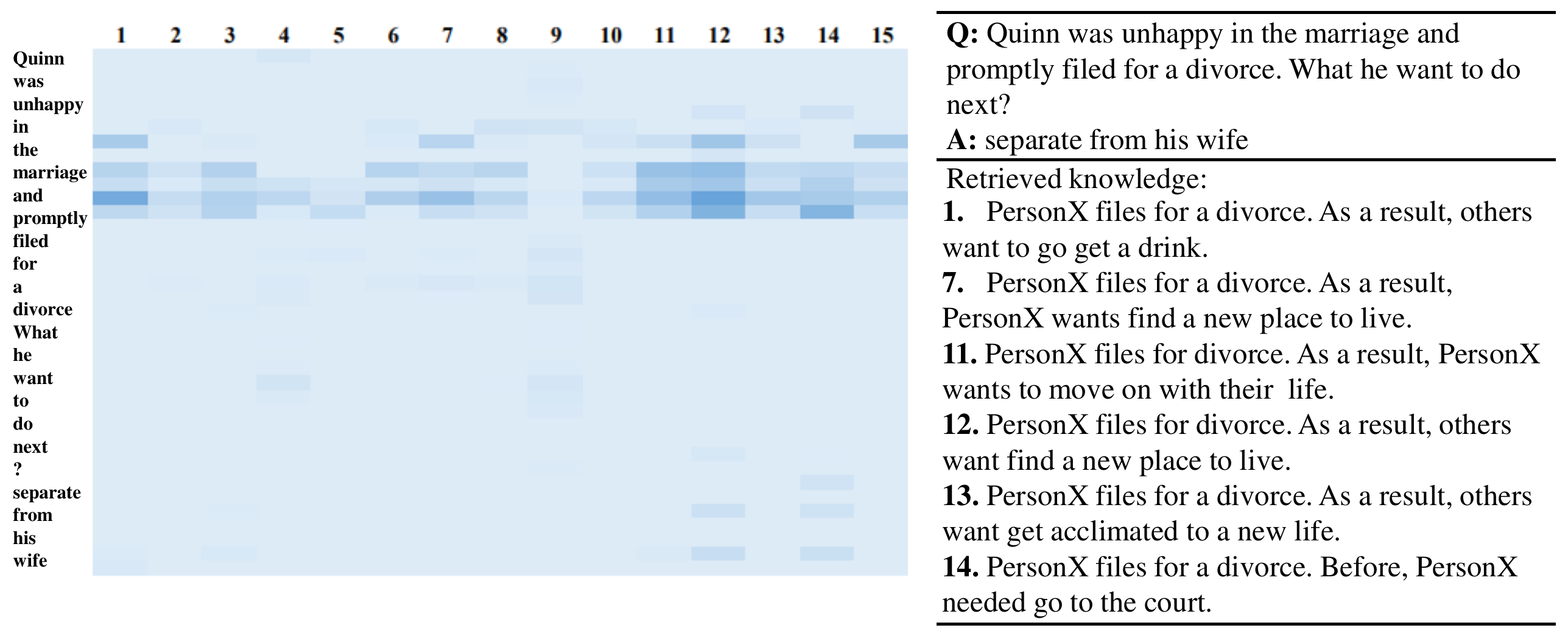}
   \caption{
   One case in the Social IQA task. The question is ``Quinn was unhappy in the marriage and promptly filed for a divorce. What does he want to do next?'' and the answer is `separate from his wife.' The matrix on the left is the activation function's output. The matrix's size is (seq\_length $\times$  \# of knowledge). Each row of the matrix represents each token in the sequence, and each column corresponds to one knowledge text we retrieved. In this experiment, we choose the top 15. 
   On the right, we list the knowledge with higher activation output.}
   \label{fig:case}
\end{figure*}

According to these results, we discover that adding the knowledge into the top 3 layers' feed-forward layers usually performs well, which supports the findings of previous work. The top three layers perform best on the Social IQA task, with a score of 72.82 (+0.56 Acc) and also achieve good results on the MedQA task (33.03 in dev and 30.01 in test). The influence of the intermediate layers, on the other hand, is not that consistent and requires more research in the future.

\subsection{Interpretability}
To study how adding the knowledge into the feed-forward layer helps our model learn from external knowledge, we select one case in which RoBERTa predicts wrong, and Kformer predicts right from the Social IQA task. 
According to \cite{dai2021knowledge}, here we use the output of the activation function to see what knowledge the model activates and uses.

As shown in Figure~\ref{fig:case}, the question is, ``Quinn was unhappy in the marriage and promptly filed for a divorce. What does he want to do next?'' and the answer is `separate from his wife.' RoBERTa chose the wrong answer, 'get himself a lawyer,' and Kformer chose the right one. Some knowledge we retrieved is useful to help the model, such as ``PersonX files for a divorce. As a result, PersonX wants to find a new place to live.'' ``PersonX files for divorce. As a result, PersonX wants to move on with their life.''. According to the retrieved knowledge, we know that Quinn would leave his ex-wife and start a new life after his divorce while finding the lawyer is the work he needed to do before. 

We can find the model mainly focuses on the 1, 7, 11, 12, and 14th knowledge it retrieved before. We list the corresponding knowledge on the right side of the Figure. As can be seen from the Figure, although the knowledge did not explicitly mention separating from wife, other information like ``acclimated to a new life'' and ``find a new place to live'' indicates the separation between the two people. The model reason over the knowledge to answer the question correctly.

\section{Related Work}

\paragraph{Knowledge Fusion.}

External knowledge can be categorized as structured knowledge and non-structured knowledge. Most works choose to use the structured knowledge such as triples or paths extracted from the knowledge graph~\cite{wang2020connecting,lv2020graph}. Besides, since the structured knowledge is often limited to some specific domains, some works also try to extract related contexts from open-domain corpora, such as Wikipedia. They use information retrieval or other text-matching technologies~\cite{lv2020graph,xu2020fusing} to extract knowledge. Our method is not limited to the type of knowledge. 
We conduct our experiments on both structured knowledge and non-structured knowledge.

\paragraph{Knowledge Fusion in Pre-trained Models.}
In terms of explicitly fusing knowledge into pre-trained models, one way is to integrate the question text and the knowledge texts as the model's input and then send it into the pre-trained model to obtain the final results. 
K-BERT \cite{liu2020k} build a sentence tree using the questions and triples in the knowledge graph and convert it into an embedding representation and a visible matrix before sending it to the encoder. 
\cite{xu2020fusing} just concatenate the knowledge into the input and directly send it to the encoder. Due to the limitation of sequence length, these methods cannot concatenate much knowledge into their models. 
Another way \cite{chang-etal-2020-incorporating,Wang2019ImprovingNL,DBLP:conf/ijcai/ZhangDCCZZC21} is to obtain the context representation ({$\mathbf{c}$}) using the pre-tained model and the knowledge representation ({$\mathbf{k}$}) through a knowledge based model. 
Then it projects the knowledge to the pre-trained model’s hidden representation space and interacts $\mathbf{k}$ and $\mathbf{c}$ via an attention network. 
The results are predicted based on the concatenation of $\mathbf{c}$ and $\mathbf{k}$ by an inference model. 

To obtain the knowledge representation, \cite{wang2020connecting} computes the knowledge embedding using a Relational Network to aggregate the retrieved paths while \cite{chang-etal-2020-incorporating} encodes the knowledge by retrofitting the pre-trained word embedding. 
These two ways may run into the Knowledge Noise problem, which occurs when too much knowledge is incorporated into a statement, causing it to lose its intended meaning. 
In our paper, inspired by previous work \cite{dai2021knowledge,geva2020transformer} on the feed-forward layers, we propose a novel way to filter and incorporate external knowledge through the feed-forward layers in Transformer. 

\section{Conclusion and Future Work}
In this paper, we investigate the feed-forward layer in Transformer and discover that infusing external knowledge into the language model through the feed-forward layer can benefit the performance. 
We propose a simple-yet-effective knowledge injection model, Kformer, which is pluggable to any pre-trained language models and adaptable to different knowledge sources.
We conduct our experiments on two multi-choice tasks (Social IQA and Med-USMLE), and Kformer achieves absolute improvements on the two tasks. 

In the future, we would integrate our fusion method during pre-training to get better knowledge representation and explore more downstream tasks to improve the generalization ability of Kformer.
Besides, it is also interesting to investigate the fundamental mechanism of knowledge storage in pre-trained language models and develop efficient models for knowledge-intensive tasks. 

\section*{Acknowledgements}
We  want to express gratitude to the anonymous reviewers for their hard work and kind comments. This work is funded by NSFC U19B2027/91846204, National Key R\&D Program of China (Funding No.SQ2018YFC000004), Zhejiang Provincial Natural Science Foundation of China (No. LGG22F030011), Ningbo Natural Science Foundation (2021J190), and Yongjiang Talent Introduction Programme (2021A-156-G).


\bibliographystyle{splncs04}
\bibliography{custom}

\end{document}